\documentclass[sigconf]{acmart}

\usepackage{xcolor}
\usepackage{multirow}
\usepackage{booktabs}
\usepackage{ulem}
\usepackage{graphicx}
\usepackage{float}
\usepackage{bbding}
\usepackage{rotating}
\usepackage{subcaption}

\AtBeginDocument{%
  }

\copyrightyear{2025}
\acmYear{2025}
\setcopyright{acmlicensed}\acmConference[MM '25]{Proceedings of the 33rd ACM International Conference on Multimedia}{October 27--31, 2025}{Dublin, Ireland}
\acmBooktitle{Proceedings of the 33rd ACM International Conference on Multimedia (MM '25), October 27--31, 2025, Dublin, Ireland}
\acmDOI{10.1145/3746027.3762026}
\acmISBN{979-8-4007-2035-2/2025/10}

\begin{document}

\title{Boosting Micro-Expression Analysis via Prior-Guided Video-Level Regression}

\author{Zizheng Guo}
\email{guozizheng@xs.ustb.edu.cn}
\orcid{0009-0005-0319-8344}
\affiliation{%
  \institution{University of Science and Technology Beijing}
  \state{Beijing}
  \country{China}
}

\author{Bochao Zou}
\authornotemark[1]
\orcid{0000-0002-2126-8159}
\email{zoubochao@ustb.edu.cn}
\affiliation{%
  \institution{University of Science and Technology Beijing}
  \state{Beijing}
  \country{China}
}

\author{Yinuo Jia}
\email{m202521009@xs.ustb.edu.cn}
\affiliation{%
  \institution{University of Science and Technology Beijing}
  \state{Beijing}
  \country{China}
}

\author{Xiangyu Li}
\email{d202510489@xs.ustb.edu.cn}
\affiliation{%
  \institution{University of Science and Technology Beijing}
  \state{Beijing}
  \country{China}
}

\author{Huimin Ma}
\orcid{0000-0001-5383-5667}
\email{mhmpub@ustb.edu.cn}
\affiliation{%
  \institution{University of Science and Technology Beijing}
  \state{Beijing}
  \country{China}
}

\renewcommand{\shortauthors}{Zizheng Guo et al.}

\begin{abstract}
Micro-expressions~(MEs) are involuntary, low-intensity, and short-duration facial expressions that often reveal an individual's genuine thoughts and emotions. Most existing ME analysis methods rely on window-level classification with fixed window sizes and hard decisions, which limits their ability to capture the complex temporal dynamics of MEs. Although recent approaches have adopted video-level regression frameworks to address some of these challenges, interval decoding still depends on manually predefined, window-based methods, leaving the issue only partially mitigated. In this paper, we propose a prior-guided video-level regression method for ME analysis. We introduce a scalable interval selection strategy that comprehensively considers the temporal evolution, duration, and class distribution characteristics of MEs, enabling precise spotting of the onset, apex, and offset phases. In addition, we introduce a synergistic optimization framework, in which the spotting and recognition tasks share parameters except for the classification heads. This fully exploits complementary information, makes more efficient use of limited data, and enhances the model's capability. Extensive experiments on multiple benchmark datasets demonstrate the state-of-the-art performance of our method, with an STRS of 0.0562 on CAS(ME)$^3$ and 0.2000 on SAMMLV. The code is available at https://github.com/zizheng-guo/BoostingVRME.
\end{abstract}




\maketitle

\section{Introduction}
\label{sec:intro}

\begin{figure*}[t] 
  \setlength{\abovecaptionskip}{2pt}
  \centering 
  \includegraphics[width=0.94\linewidth]{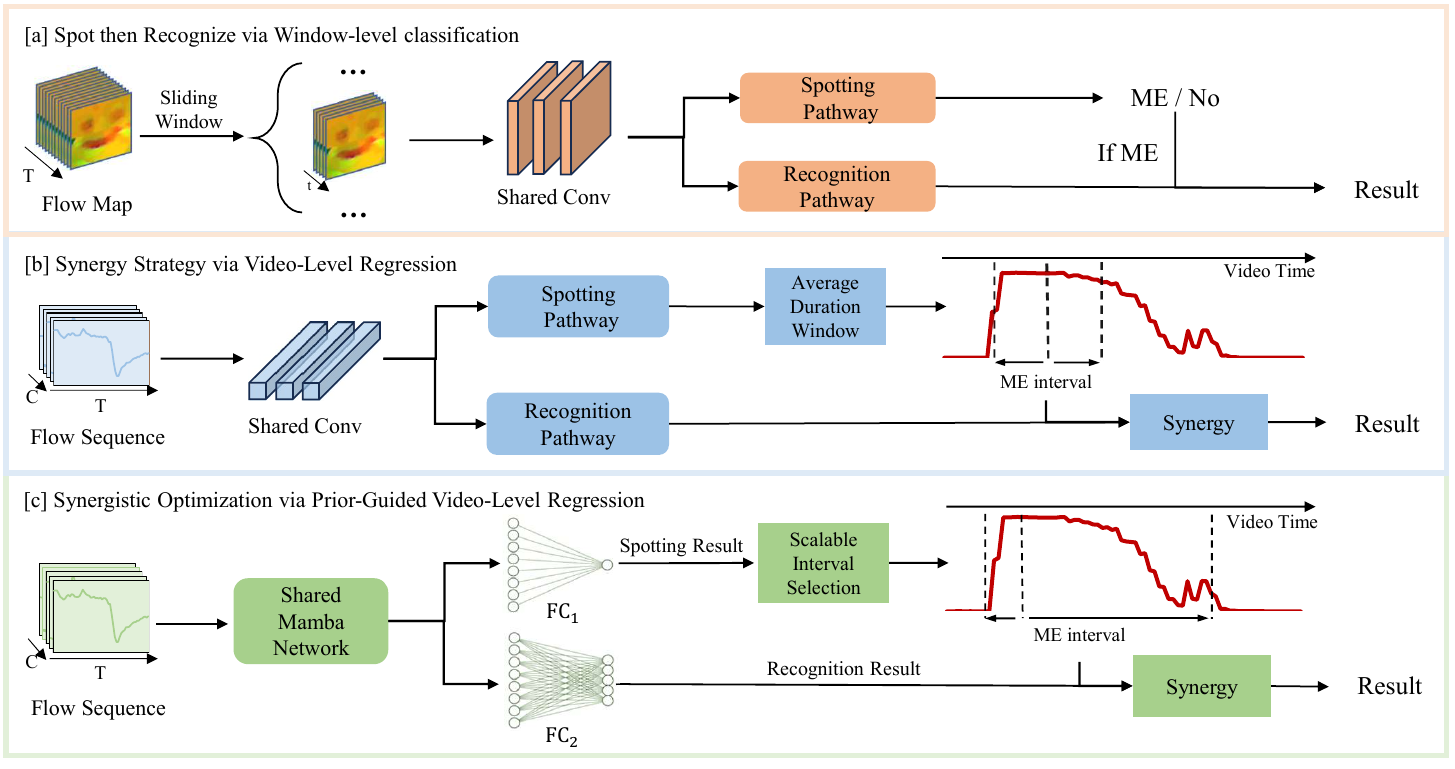}  
  \caption{(a) Previous methods with window-level classification. (b) Synergy strategy via video-level regression. (c) The proposed synergistic optimization via prior-guided video-level regression, in which all parameters are shared between spotting and recognition, except for the classification heads, and decoding is performed via a prior-guided scalable interval selection strategy.}
  \label{fig:1}
\end{figure*}

Facial expressions, as an important medium for conveying individual emotions, are typically divided into macro-expressions (MaEs) and micro-expressions (MEs). In comparison to MaEs, which can be consciously controlled and concealed, MEs are characterized by their extremely short duration and low intensity, and are often produced unconsciously. MEs are more likely to be elicited when individuals attempt to partially hide or suppress their genuine emotions, thereby inadvertently revealing subtle information such as true intentions, choices, and preferences. As a result, ME analysis holds significant value in high-risk scenarios such as medical diagnosis, public security, criminal investigation, and political or business negotiations~\cite{tpami-review, tac-survey}.

ME analysis requires the precise spotting and recognition of MEs within long videos. Previous research typically divides this task into two fundamental subtasks: spotting and recognition. Both of these subtasks have experienced a methodological transition from traditional signal processing techniques to deep learning approaches~\cite{casme3}. However, due to the low signal-to-noise ratio of MEs in long videos, traditional methods still exhibit strong competitiveness in the spotting task~\cite{MEGC2023}, whereas deep learning-based methods have achieved significant advances in the recognition task~\cite{recog-tac24, recog_cvpr23, mol}.

Despite considerable progress in both of these subtasks, research on comprehensive ME analysis remains limited, although it is more aligned with practical application needs. Currently, exploration in this field is still at an early stage. Some classical methods (e.g.,~\cite{li2017},~\cite{Liong2017_MA}) adopt handcrafted features for both spotting and recognition, but they are mainly limited to the analysis of single, short ME video clips, with little consideration of interference from MaEs. In recent years, methods such as MEAN~\cite{MEAN}, SFAMNet~\cite{sfamnet}, and subsequent research~\cite{megc24-1, megc24-2} have leveraged deep learning models to sequentially perform spotting and recognition in long videos. However, these approaches often rely on fixed-size sliding windows and hard classifiers, making it difficult to fully capture the diverse temporal dynamics of MEs. ME-TST~\cite{METST, metst+} has achieved synergistic spotting and recognition via the temporal state transition architecture. However, its interval decoding phase still relies on hand-crafted strategies that involve preset window lengths. These strategies impose artificial temporal constraints and diminish the unique advantages of the architecture, thereby limiting its overall effectiveness. Furthermore, most current methods process the two tasks in separate stages or only share weights at low-level feature layers, without fully exploring the synergistic potential between the spotting and recognition tasks.

To address these challenges, this paper proposes a prior-guided video-level regression to boost ME analysis. The proposed method introduces a scalable interval selection strategy that comprehensively accounts for the temporal evolution, duration, and class distribution characteristics of MEs, achieving precise spotting of the onset, apex, and offset phases through a hierarchical thresholding mechanism. In addition, this work proposes a synergistic optimization framework that enables parameter sharing between the spotting and recognition tasks, except for the classification heads. This approach further enhances the inter-task synergy and facilitates the mining of complementary information, thereby improving the model's learning capabilities. This approach not only makes effective use of limited data resources but also reduces the overall number of model parameters and computational cost.

The main contributions can be summarized as follows:

$\bullet$ We propose a prior-guided video-level regression framework for ME analysis. A scalable interval selection strategy guided by priors is introduced to fully leverage the strengths of video-level regression, enabling accurate spotting of ME onset, apex, and offset.

$\bullet$ We introduce a synergistic optimization framework that leverages complementary information between spotting and recognition tasks to enhance the model's learning capacity and maximize the utilization of limited data resources.

$\bullet$ Extensive experiments on multiple benchmarks demonstrate that the proposed method achieves state-of-the-art performance.
\section{Related Work}
\label{sec:relatedwork}

\subsection{Micro-Expression}
\textbf{ME Spotting:}
The overall development of ME spotting has demonstrated a shift from signal processing methods based on handcrafted features to deep learning approaches. Early research primarily relied on handcrafted features such as optical flow~\cite{review74,review75,review76,review77}, Local Binary Patterns (LBP)~\cite{review79}, Histogram of Oriented Gradients (HOG)~\cite{cvpr25-4}, and Scale Invariant Feature Transform (SIFT)~\cite{review80}. However, these traditional methods are often limited by their sensitivity to noise, high computational complexity, and strong dependence on parameter tuning~\cite{tpami-review}. With the advent of machine learning, data-driven approaches have become mainstream. For example, some studies have utilized Support Vector Machine (SVM) classifiers~\cite{casme38,casme42}. Many methods incorporate handcrafted features as prior knowledge into deep architectures, leveraging various network structures (e.g., multi-stream)~\cite{lssnet,softnet} or spatio-temporal models (e.g., 3D-CNNs)~\cite{3D-CNN,mesnet}. While many approaches spot ME intervals based on keyframes~\cite{casme39,weakly36,weakly37}, for extremely subtle MEs, dynamic temporal information tends to be significantly more informative than static information from a single frame. Recently, interval-based methods~\cite{weakly40,weakly41,weakly42,lssnet,lgsnet} have achieved significant progress by incorporating temporal information. Nevertheless, deep learning approaches have not shown a clear advantage over traditional methods in the ME spotting task~\cite{MEGC2023}.

\textbf{ME Recognition:}
Early methodologies for ME recognition predominantly utilized handcrafted features—such as Local Binary Patterns from Three Orthogonal Planes (LBP-TOP)~\cite{tpami25-7,cvpr25-26} and Histogram of Oriented Gradients (HOG)~\cite{cvpr25-4}—to capture subtle facial muscular movements. Subsequently, optical flow-based descriptors like Main Directional Mean Optical Flow (MDMO)~\cite{mdmo} and Fuzzy Histograms of Optical Flow (FHOOF)~\cite{review87,tpami25-16} were developed to enrich motion representations. Nevertheless, these conventional approaches tend to be susceptible to noise and are heavily reliant on the precise localization of keyframes, thereby limiting their robustness and practical utility. The evolution of machine learning has catalyzed a shift towards data-driven methods, with early approaches applying classifiers like Support Vector Machines (SVM)~\cite{review71,review84,mdmo} and K-Nearest Neighbors (KNN)~\cite{review87,review92} to handcrafted features.
More recently, deep learning models have enabled the automatic extraction of highly informative spatio-temporal features. For example, architectures such as OFF-ApexNet~\cite{off-apexnet}, STSTNet~\cite{ststnet}, and GLEFFN~\cite{cvpr25-10} integrate optical flow to capture both global and local motion patterns. Further research has advanced ME recognition through improved feature extraction and temporal modeling techniques~\cite{rcnn,cmnet,selfme,frldgt,soda4mer,mol}. Notwithstanding this progress, the majority of current ME recognition methods still depend on pre-defined ME intervals and precise keyframe annotations, which limits their applicability in real-world scenarios.

\textbf{ME Analysis:}
Compared to the substantial progress in ME spotting and recognition as separate tasks, few studies have directly addressed the integrated task of ME analysis, despite its greater relevance to real-world application scenarios. Currently, advances in this domain remain limited. Traditional signal processing methods~\cite{li2017,Liong2017_MA} rely on handcrafted features for both spotting and recognition. However, these approaches are typically designed for short video clips containing a single ME and do not consider the presence of MaEs.
More recent deep learning approaches, including MEAN~\cite{MEAN} and SFAMNet~\cite{sfamnet}, sequentially perform spotting and recognition in long video sequences. In MEGC2024~\cite{megc2024}, USTC-IAT-United~\cite{megc24-1} introduced the VideoMAE framework and multi-scale strategies to more effectively capture the dynamics of MEs. Meanwhile, He et al.~\cite{megc24-2} mitigated the issue of data imbalance through a tailored data preparation strategy and employed a three-stage pipeline to enhance performance. ME-TST~\cite{METST,metst+} achieved synergistic spotting and recognition by leveraging a temporal state transition architecture.
Nevertheless, these previous methods remain constrained by their reliance on average temporal windows and have not fully explored the relationship between spotting and recognition. As a result, a truly unified and adaptive solution for integrated ME analysis in unconstrained scenarios is still lacking.

\subsection{Mamba}
The recently proposed Mamba architecture has garnered significant attention due to its data-dependent State Space Models (SSMs) and parallel selection mechanism~\cite{mamba,mamba2}. This design achieves linear computational complexity in sequence modeling, enabling both efficient processing of long sequences and effective modeling of long-range dependencies compared to transformer-based models~\cite{transformer,vivit}. Building on this foundation, subsequent works~\cite{visionmamba,videomamba,rhythmmamba} have extended Mamba to a variety of visual tasks.
In the field of ME analysis, ME-TST~\cite{METST,metst+} was the first to utilize Mamba for this task by mapping the evolution of ME dynamics onto the hidden states of a state space model. However, ME-TST still relies on a preset window-based interval decoding approach, thereby preventing the theoretical potential of state space models from being fully realized in ME analysis.
\section{Methodology}
\label{sec/3_methodology} 

\subsection{The General Framework} 
The framework of the proposed method is illustrated in Fig.~\ref{fig:1}(c). Following the design in~\cite{metst+}, for a given facial video input, optical flow is extracted using overlapping sliding windows, which are then aggregated to form the full-video optical flow. Optical flow sequences are extracted from key facial regions of interest (ROIs) at multiple granularities, using the first frame of each window as the reference. For each ROI and each granularity, the extracted sequence is treated as a separate channel, resulting in $X_{input} \in \mathbb{R}^{T\times C_1}$, where $T$ denotes the length of the temporal segment and $C_1$ is the total number of channels. This input is then fed into a neural network based on the SlowFast Mamba architecture~\cite{slowfast,mamba}, which is designed to effectively explore the relationship between ROI motion combinations and MEs, yielding ME features $X_{feature} \in \mathbb{R}^{T\times C_2}$, where $C_2$ is the feature dimension of the neural network. Finally, these features are respectively input into the spotting predictor and recognition predictor to produce the spotting result $X_{spotting} \in \mathbb{R}^{T\times 1}$ and the recognition result $X_{recognition} \in \mathbb{R}^{T\times N}$, where $N$ denotes the number of emotion categories plus one (i.e., the neutral class). Both $X_{spotting}$ and $X_{recognition}$ can be interpreted as probability distributions. In the post-processing stage, the scalable interval se+lection strategy is employed, utilizing hierarchical thresholding to determine the ME intervals.

\subsection{Scalable Interval Selection Strategy}
Compared to traditional window-level classification frameworks, the core advantage of the video-level regression paradigm lies in its ability to overcome the dual limitations of fixed temporal window size and hard decision classification. Specifically, conventional approaches commonly rely on preset sliding windows, which impose inflexible temporal boundaries. Such fixed windows are inherently ill-suited to MEs, which exhibit pronounced heterogeneity in their temporal dynamics, including onset, apex, offset, and duration. Additionally, hard classification produces discrete outputs, failing to capture the continuous intensity changes of MEs. Although recent research has adopted a video-level regression framework, the final interval decoding stage still adheres to the window-level classification paradigm, where potential event points are identified using peak detection algorithms and intervals are defined as fixed windows centered on these event points. This strategy remains fundamentally constrained by hand-crafted temporal assumptions, thereby preventing the full exploitation of the theoretical advantages of video-level regression models in temporal modeling.

To address this bottleneck, we introduce a scalable interval selection strategy informed by the intrinsic temporal priors of MEs. As illustrated in Fig.~\ref{fig:1}, compared to the fixed average duration window approach (b), the proposed method (c) introduces greater flexibility. When a peak is detected as a potential event onset, approach (b) selects a window centered at the event point as the proposal, with a fixed width equal to the average ME duration range k. In contrast, the proposed method (c) performs bidirectional extension using a two-level threshold mechanism and reselects the peak point upon encountering additional peaks within the extended region. Specifically, within the range k, a lower threshold is employed, and a position is only discarded if the probability falls below this threshold in two consecutive frames. Beyond the range k, a higher threshold is adopted to robustly extend the potential duration of detected MEs, terminating detection only when the probability remains below this high threshold for two consecutive frames. This dual-threshold mechanism effectively balances the precise detection of local abrupt features with the maintenance of overall temporal continuity, thereby enabling continuous modeling of the entire ME lifecycle—including onset, apex, and offset phases—and facilitating the adaptive capture of MEs with varying durations.

\subsection{Synergistic Optimization}
As shown in Fig.~\ref{fig:1}, previous works~\cite{MEAN,METST,metst+} have been limited to sharing only a preliminary feature extraction module. In contrast, this paper further models the tasks of spotting and recognition in ME analysis as a unified synergistic optimization framework. Except for the task-specific prediction heads (final fully connected layers), all other network layers share parameters for feature extraction. The architectural design is based on the critical assumption regarding the essential nature of the two tasks. Unlike most multi-objective optimization tasks that are typically conflicting, these two tasks, despite their significant differences in high-level semantic representations, share a common foundation. Specifically, the spotting task centers on detecting minute motion changes in specific facial regions, while the recognition task focuses on the combination of movements within these regions. However, in the feature space, both tasks rely on the effective capture of motion information in ROIs on the face. This synergistic optimization framework not only reduces the overall model's parameter count and computational overhead, thereby improving computational efficiency, but also enhances the model’s learning capability by leveraging complementary information between the tasks. This is particularly valuable in scenarios like ME analysis, where data are scarce and modeling each task in isolation would fail to fully utilize the limited available samples.

Furthermore, the ME analysis task faces two severe class imbalance issues: first, the number of non-ME segments is substantially greater than that of ME segments; second, the number of samples from the negative emotion category vastly exceeds those from other emotion categories. To mitigate this problem, during training, we randomly sample non-ME segments at a ratio based on the number of ME segments and assign loss weights according to the distribution of emotion categories. Additionally, during inference, we introduce probability penalization to suppress the predicted scores for the majority classes and refine the strategy for emotion determination within ME intervals in the post-processing stage, ensuring that even subtle signals from minority classes are prioritized and classified accordingly.
\label{sec/4_experiment}
\section{Experiment}

\begin{table}[t]
  \centering
  \setlength{\tabcolsep}{6.5pt}
    \begin{tabular}{cccc}
    \toprule
    Type & Method & CAS(ME)$^3$ & SAMMLV \\
    \midrule
    \multirow{14}[2]{*}{S} 
          & MDMD \cite{he}  & -     & 0.0364 \\
          & SP-FD \cite{zhang2020} & 0.0103 & 0.1331 \\
          & OF-FD \cite{he-megc2021-1} & 0.0000 & 0.2162 \\
          & SOFTNet \cite{softnet}  & - & 0.1520 \\
          & LSSNet \cite{lssnet} & 0.0653 & 0.2180 \\
          & ABPN \cite{abpn} & - & 0.1689 \\
          & MTSN \cite{mtsn} & - & 0.0878 \\
          & LGSNet \cite{lgsnet} & 0.0990 & - \\
          % & Yang et al. \cite{yang2021facial} & - & 0.1155 \\
          & Li et al. \cite{li2024learning} & - & 0.2541 \\
          & WCMN \cite{zhou2024micro} & - & 0.2330 \\
          & Ofct \cite{ofct} & - & 0.2466 \\
          & MSOF \cite{msof} & - & 0.2457 \\
          & Causal-Ex \cite{Causal-Ex} & - & 0.2010 \\
          & MC-WES \cite{yu2025weakly} & 0.0000 & 0.1350 \\
    \midrule
    \multirow{6}[2]{*}{S \& R} & MEAN \cite{MEAN}  & 0.0283 & 0.0949 \\
          & SFAMNet \cite{sfamnet}& 0.0716 & - \\
          & ME-TST \cite{METST}  & 0.0802 & 0.2167 \\
          & He et al. \cite{megc24-2}& - & 0.1900 \\
          & USTC-IAT-United \cite{megc24-1}& - & 0.1800 \\
          & Ours  & \textbf{0.0997} & \textbf{0.2906} \\
    \bottomrule
    \end{tabular}%
    \caption{Performance Evaluation of ME Spotting: F1 Score.}
  \label{tab:mes}%
\end{table}%

\begin{table}[t]
  \vspace{-0.6cm}
  \centering
  \setlength{\tabcolsep}{14.0pt}
    \begin{tabular}{cccc}
    \toprule
    Type & Method & UF1$\uparrow$  & UAR$\uparrow$\\
    \midrule
    \multirow{4}[2]{*}{R} & STSTNet \cite{ststnet} & 0.3795 & 0.3792 \\
          & RCN-A \cite{RCN-A} & 0.3928 & 0.3893 \\
          & FeatRef \cite{FR} & 0.3493 & 0.3413 \\
          & AlexNet \cite{casme3} & 0.3001 & 0.2982 \\
    \midrule
    \multirow{4}[2]{*}{S \& R} & MEAN \cite{MEAN} & 0.3894 & 0.4004 \\
          & SFAMNet \cite{sfamnet} & 0.4462 & 0.4767 \\
          & ME-TST \cite{METST} & 0.4754 & 0.4878 \\
          & Ours  & \textbf{0.5638} & \textbf{0.5524} \\
    \bottomrule
    \end{tabular}%
  \caption{Performance Evaluation of ME Recognition on CAS(ME)$^3$.}
  \label{tab:mer}%
  \vspace{-0.4cm}
\end{table}%

\begin{table*}[t]
  \centering
  \setlength{\tabcolsep}{13.5pt}
    \begin{tabular}{ccccccc}
    \toprule
    \multirow{2}[4]{*}{Method} & \multicolumn{3}{c}{CAS(ME)$^3$} & \multicolumn{3}{c}{SAMMLV} \\
    \cmidrule{2-7}   & Analysis & Spotting  & Recognition & Analysis & Spotting  & Recognition \\
    \midrule
    MEAN \cite{MEAN} & 0.0100 & 0.0283 & 0.3532 & 0.0499 & 0.0949 & 0.5263 \\
    SFAMNet \cite{sfamnet} & 0.0331 & 0.0716 & 0.4619 & -     & -     & - \\
    ME-TST \cite{METST}  & 0.0387 & 0.0802 & 0.4830 & 0.1356 & 0.2167 & 0.6259 \\
    He et al. \cite{megc24-2}& -     & -     & - & 0.0900 & 0.1900  & 0.5100 \\
    USTC-IAT-United \cite{megc24-1}& -     & -     & - & 0.1100 & 0.1800  & 0.5800 \\
    Ours  & \textbf{0.0562} & \textbf{0.0997} & \textbf{0.5638} & \textbf{0.2000} & \textbf{0.2906} & \textbf{0.6882} \\
    \bottomrule
    \end{tabular}%
  \caption{Performance Evaluation of ME Analysis: STRS for Analysis, F1 Score for Spotting and Recognition.}
  \label{tab:mea}%
\end{table*}%

\begin{table*}[t]
  \vspace{-0.5cm}
  \centering
  \setlength{\tabcolsep}{10.8pt}
    \begin{tabular}{cccccccc}
    \toprule
    \multirow{2}[4]{*}{Method} & \multicolumn{1}{c}{} & \multicolumn{3}{c}{CAS(ME)$^3$ Unseen Subset} & \multicolumn{3}{c}{SAMMLV Unseen Subset} \\
    \cmidrule{3-5} \cmidrule{6-8}
        & Overall Analysis & Analysis & Spotting & Recognition & Analysis & Spotting & Recognition \\
    \midrule
    Ours & \textbf{0.09} & \textbf{0.06} & \textbf{0.10}  & \textbf{0.65} & \textbf{0.06} & 0.09 & \textbf{0.67} \\
    USTC-IAT & 0.06 & 0.04 & 0.07 & \textbf{0.65} & \textbf{0.06} & \textbf{0.12} & 0.47 \\
    Gormanv & 0.05  & 0.02 & 0.06 & 0.28 & 0.04 & 0.07 & 0.62 \\
    Li & 0.03  & 0.02 & 0.03 & 0.61 & 0.03 & 0.06 & 0.61 \\
    MEAN & -  & 0.01 & 0.04 & 0.33 & 0.01 & 0.03 & 0.22 \\
    \bottomrule
    \end{tabular}%
  \caption{Leaderboard of the MEGC2025-testSet: STRS for Analysis, F1 Score for Spotting and Recognition.}
  \label{tab:performance_comparison}%
\end{table*}%

\begin{table*}[t]
  \vspace{-0.5cm}
  \centering
  \setlength{\tabcolsep}{6pt}
  \begin{tabular}{cccccccccccc}
    \toprule
    \multirow{2}{*}{SISS} & \multirow{2}{*}{SO} & \multicolumn{5}{c}{CAS(ME)$^3$} & \multicolumn{5}{c}{SAMMLV} \\
    \cmidrule(lr){3-7} \cmidrule(lr){8-12}
    & & Analysis & Spotting & Recognition & $IoU_{tp}$ & $IoU_{all}$ & Analysis & Spotting & Recognition & $IoU_{tp}$ & $IoU_{all}$ \\
    \midrule
    × & × & 0.0487 & 0.0912 & 0.5338 & 0.6827 & 0.4221 & 0.1848 & 0.2723 & 0.6787 & 0.7354 & 0.6073 \\ 
    × & \checkmark & 0.0480 & 0.0901 & 0.5327 & 0.6713 & 0.4426 & 0.1989 & \textbf{0.2943} & 0.6760 & 0.7340 & 0.6242  \\ 
    \checkmark & \checkmark & \textbf{0.0562} & \textbf{0.0997} & \textbf{0.5638} & \textbf{0.7039} & \textbf{0.4750}  & \textbf{0.2000} & 0.2906 & \textbf{0.6882} & \textbf{0.7381} & \textbf{0.6308}  \\
    \bottomrule
  \end{tabular}%
  \caption{Ablation study: STRS for Analysis, F1 Score for Spotting and Recognition.}
  \label{tab:ablation}%
\end{table*}
    
\subsection{Datasets and Evaluation Metrics}
\textbf{Datasets.} Three widely used long-video datasets are employed: CAS(ME)$^3$~\cite{casme3}, SAMMLV~\cite{samm,SAMMLV}, and MEGC2025-testSet~\cite{megc2025,MEGC2023}.
\textbf{CAS(ME)$^3$} comprises 1,300 videos collected from 100 subjects at a frame rate of 30 fps. Annotations are provided for 3,342 MaEs and 860 MEs.
\textbf{SAMMLV} includes 147 videos collected from 32 participants at a frame rate of 200 fps. Annotations are provided for 343 MaEs and 159 MEs.
\textbf{MEGC2025-testSet}~\cite{megc2025,MEGC2023} serves as the official evaluation benchmark for the MEGC2025 challenge. This test set consists of 30 long videos, including 10 from the SAMM dataset and 20 from CAS(ME)$^3$. Frame rates are consistent with those of the respective source datasets~(i.e., 200 fps and 30 fps). 

\textbf{Evaluation metrics.} For spotting, we report precision, recall, and F1-score. For recognition, we utilize the F1-score, unweighted average recall (UAR)~\cite{uf1uar}, and unweighted F1-score (UF1). Overall performance is quantified by the Spot-Then-Recognize Score (STRS)~\cite{MEAN}. The spotting evaluation is based on the Intersection-over-Union (IoU) metric~\cite{iou}. A predicted interval is regarded as a True Positive (TP) when its IoU value with the corresponding ground truth interval is greater than 0.5. Furthermore, to enable a more fine-grained assessment of onset and offset boundary accuracy, we introduce two additional metrics, $IoU_{tp}$ and $IoU_{all}$. These are calculated as follows: ${IoU}_{tp} = avgIoU(IoU>0.5), {IoU}_{all} = avgIoU(IoU>0).$ Here, $avgIoU$ denotes the average IoU calculated over all intervals that satisfy the specified threshold.

\subsection{Implementation Details}
The proposed method was implemented using PyTorch. Optical flow was computed using the Gunnar Farneback algorithm~\cite{flow}. For the spotting task, the Mean Squared Error (MSE) was adopted as the loss function, while cross-entropy loss was utilized for the recognition task. The flow sequence length was set to 50 frames, and the outputs were concatenated to produce ME probabilities across the entire video for interval selection. The model was trained for 50 epochs with a learning rate of 3e-4. All experiments were performed on an NVIDIA RTX 4090 GPU.

\begin{figure*}[t]
  \setlength{\abovecaptionskip}{3pt}
  \centering  
  \includegraphics[width=0.96\linewidth]{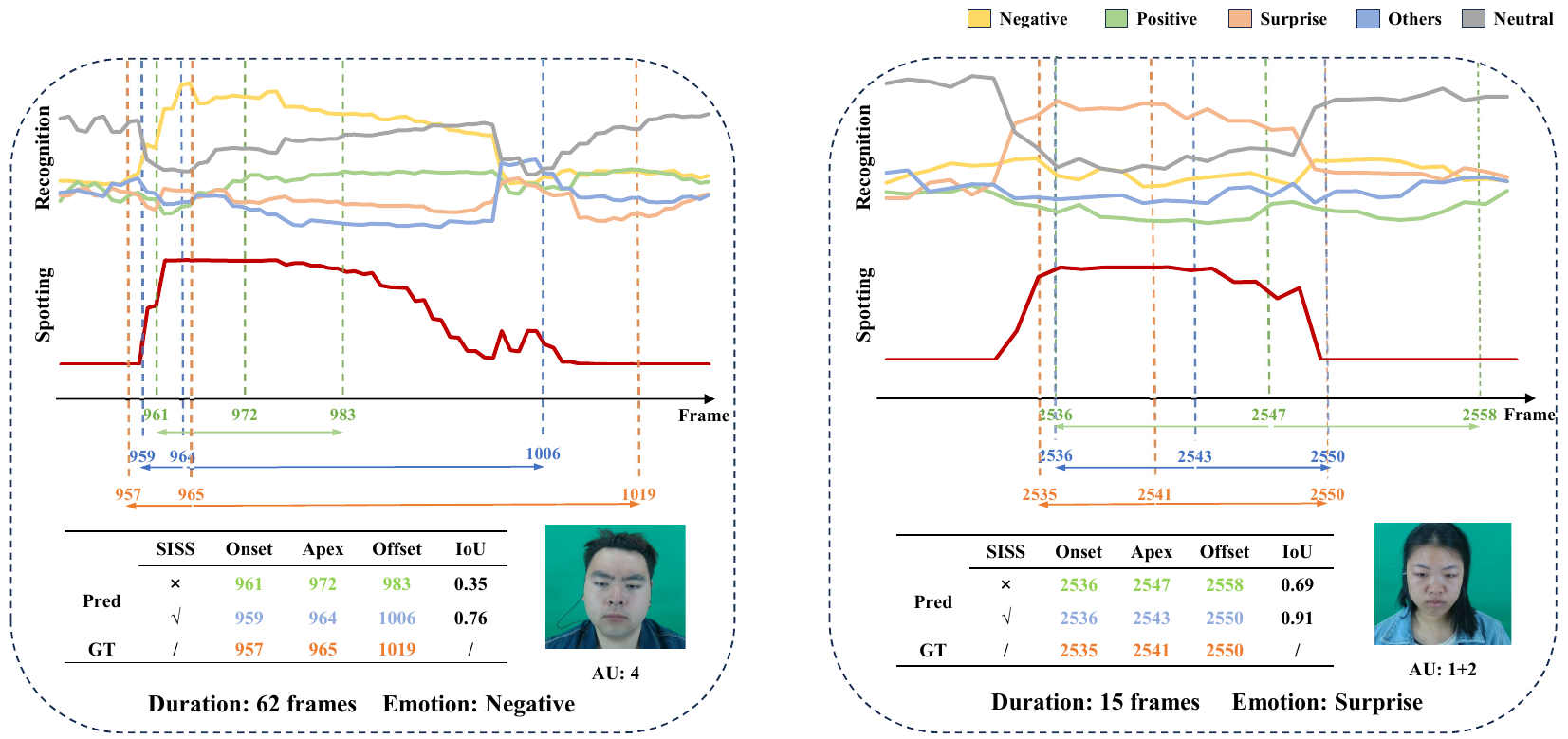}  
  \caption{Visualization of the proposed Scalable Interval Selection Strategy.}
  \label{fig:vis}
\end{figure*}

\subsection{ME Subtask Evaluation}
As shown in Table~\ref{tab:mes} and Table~\ref{tab:mer}, we compared the spotting and recognition performance of the proposed method with previous approaches, following the protocol described in~\cite{casme3,sfamnet}. These approaches include methods focused solely on spotting (type S), solely on recognition (type R), and simultaneous spotting and recognition (type S\&R). The Leave-One-Subject-Out cross-validation protocol was employed for evaluation. The results demonstrate that our method achieves state-of-the-art performance compared to other approaches that perform simultaneous spotting and recognition, and that its performance on the individual tasks of spotting and recognition also surpasses that of methods dedicated solely to spotting or recognition.

\subsection{ME Analysis Evaluation}
Following the evaluation protocol outlined in~\cite{MEAN,METST}, we performed an extensive evaluation of ME analysis to rigorously assess the performance of our approach. Table~\ref{tab:mea} presents a comparison with existing methods, where our technique achieves STRS scores of 0.2000 on the SAMMLV dataset and 0.0562 on the CAS(ME)$^3$ dataset, consistently obtaining state-of-the-art results across all three primary evaluation metrics. These results highlight the superior discriminative capability of the proposed prior-guided video-level regression architecture, particularly in capturing the subtle and transient patterns characteristic of MEs.

To further validate our approach, we conducted evaluations on the MEGC2025-testSet~\cite{megc2025}. As shown in Table~\ref{tab:performance_comparison}, our method achieved first place on the leaderboard, surpassing all existing solutions on key metrics. This further substantiates the robustness and generalizability of our approach in dealing with real-world ME analysis tasks. Nevertheless, it is also evident that the overall analysis performance is significantly constrained by the limitations of the spotting, indicating substantial room for further improvement.

\subsection{Ablation Study}
We conducted ablation studies to evaluate the impact of the Scalable Interval Selection Strategy (SISS) and Synergistic Optimization (SO). The results, as shown in Table~\ref{tab:ablation}, demonstrate that the scalable interval selection strategy substantially improves overall performance across all metrics. The improvements in $IoU_{tp}$ and $IoU_{all}$ indicate that the scalable interval selection strategy enables more accurate spotting of ME onset and offset. In addition, it can be observed that synergistic optimization further enhances the overall performance while reducing the number of parameters and computational cost, providing strong evidence for the complementary and mutually beneficial relationship between the spotting and recognition tasks.

\subsection{Visualization} 
Two examples of the results on the CAS(ME)$^3$ dataset are visualized in Fig.~\ref{fig:vis}. It can be observed that the identification of potential event points via peak detection, combined with fixed-size temporal window expansion, is essentially limited by the manually preset temporal constraints. This approach fails to fully leverage the theoretical advantages of video-level regression architectures in temporal modeling. When the duration is relatively long (left panel) or short (right panel), the fixed window is unable to adapt to the varying durations of MEs, and the detected peaks also exhibit significant deviations. In contrast, the proposed scalable interval selection strategy fully exploits the advantages of the video-level regression architecture. It accurately captures the onset, apex, and offset frames of MEs based on the probability at each time point. Moreover, it can be observed that within the occurrence interval of MEs, the corresponding emotion probability and spotting probability evolve synchronously, reflecting the complementary effects of spotting and recognition under the synergistic optimization framework. Overall, the proposed method effectively models the temporal dynamics of MEs and fully demonstrates the superiority of the prior-guided video-level regression architecture.
\label{sec/5_conclusion}
\section{Conclusion}

In this paper, we propose a prior-guided video-level regression architecture, which incorporates a scalable interval selection strategy and synergistic optimization to boost ME analysis. Our method achieves true video-level regression and attains state-of-the-art performance across multiple evaluation protocols. The proposed architecture is better aligned with the temporal characteristics of MEs and is not restricted by hand-crafted temporal constraints. Furthermore, it is more applicable to real-world scenarios, as it enables direct spotting and recognition of MEs in long video sequences. We believe this approach can serve as a new baseline for the ME community. In the future, further exploration of the complementary relationship between spotting and recognition may represent a promising direction.

\begin{acks}
This work was supported in part by the National Natural Science Foundation of China (62206015, 62227801), the Fundamental Research Funds for the Central Universities (FRF-KST-25-008), and the Young Scientist Program of The National New Energy Vehicle Technology Innovation Center (Xiamen Branch).
\end{acks}

\bibliographystyle{ACM-Reference-Format}
\bibliography{sample-base}

\end{document}